\pdfoutput=1
\documentclass{article}
\usepackage{acra}
\usepackage{graphicx}
\usepackage{booktabs}
\usepackage{amsmath}
\usepackage{xcolor}
\usepackage[font=small,labelfont=bf]{caption}
\usepackage{multirow}
\usepackage{tikz}
\usetikzlibrary{arrows.meta,calc}
\newcommand{\PaperAuthors}{Younsoo Kim$^{1}$, Sangmin Song$^{1}$, Sarath Kodagoda$^{1}$,\\{\Large\bf Marc Carmichael$^{1}$ and Karthick Thiyagarajan$^{2}$}}
\newcommand{\PaperAffiliation}{%
  {\normalfont\normalsize $^{1}$University of Technology Sydney, NSW, Australia}\\
  {\normalfont\normalsize $^{2}$Western Sydney University, NSW, Australia}}

\title{TacGELLO: Tactile Contact Feedback for a\\
       3D-Printed Teleoperation Leader}
\author{\PaperAuthors \\ \PaperAffiliation}

\begin{document}
\maketitle
\begin{abstract}
Passive leader arms transmit an operator's motion without returning remote contact or load cues. TacGELLO uses the existing servos of a 3D-printed leader for tactile contact feedback at the gripper trigger and current-based load cues at the arm joints. AnySkin contact detection activates a trigger position hold, while changes in follower joint current adjust leader-servo output limits to command resistance during lifting.

Across six UR3 sessions with one operator, sent-command tracking error was 0.15--0.36$^\circ$ RMSE, with estimated trajectory lag of 82--98\,ms. Tactile onset was logged in 60 of 62 segmented TacGELLO grasp intervals. Median commanded closure after tactile or width-based contact references was 0.7--2.1\,mm with TacGELLO and 13.0--13.9\,mm with the original interface. Separate lifting recordings showed arm-cue command activation.

Command dynamics and contact references differed between interfaces, preventing attribution of the closure difference to tactile feedback alone. 
\end{abstract}

\section{Introduction}
\label{sec:intro}
\begin{figure}[t]
\centering
\includegraphics[width=\linewidth]{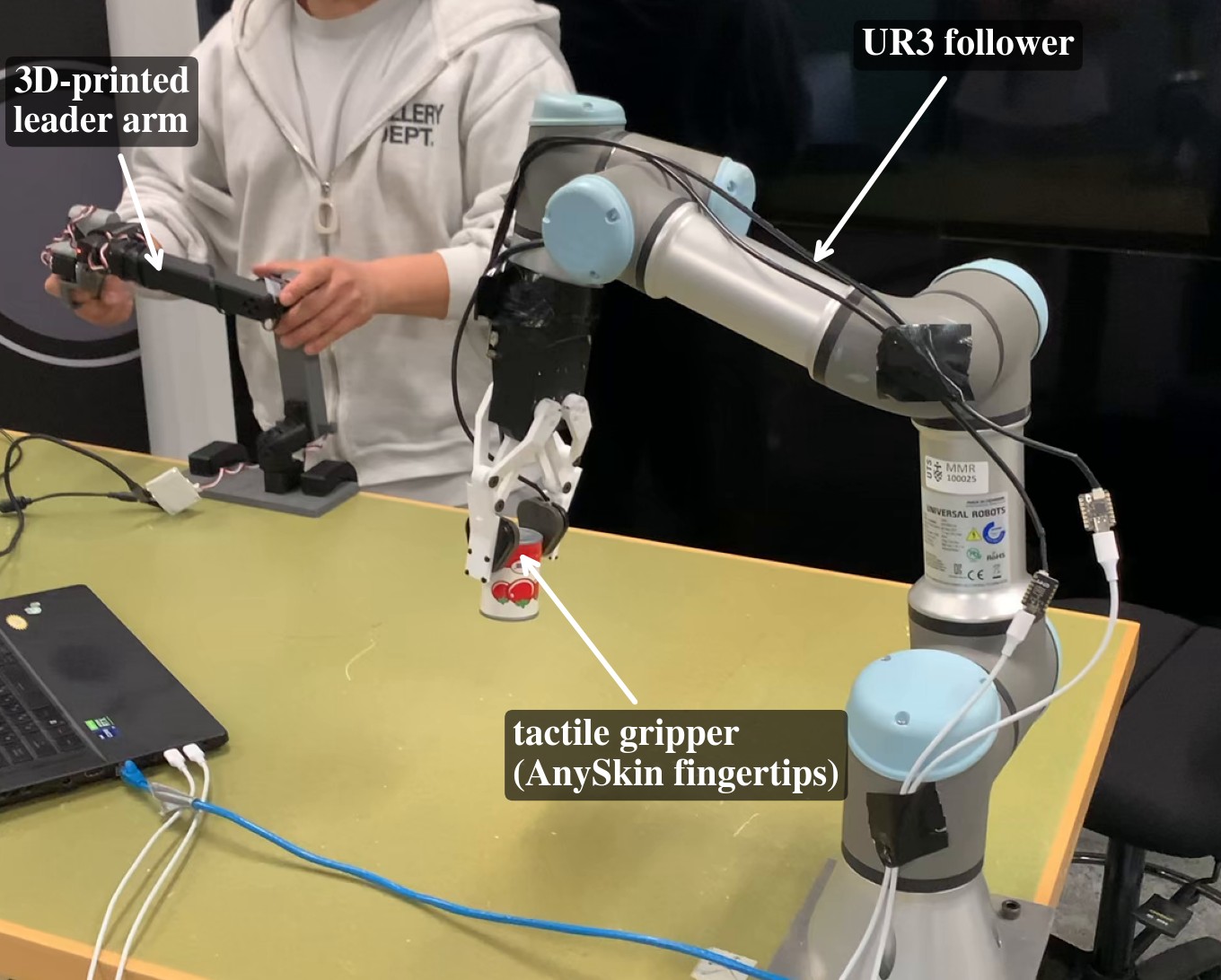}
\caption{TacGELLO in use. The operator moves the 3D-printed leader
and the UR3 follows it, here lifting a can with the tactile gripper.
When the AnySkin fingertips detect contact, the leader's trigger
servo holds the trigger, so the operator feels the grasp.}
\label{fig:overview}
\end{figure}
Leader-arm interfaces map an operator's joint motion to a robot
manipulator for teleoperation and demonstration collection.
GELLO uses a low-cost, 3D-printed replica of the follower's
kinematics~\cite{wu2024gello}. ALOHA provides a related interface
for bimanual manipulation~\cite{zhao2023aloha}.
In their passive configurations, both transmit motion without
returning remote contact forces. An operator grasping a fragile
object must infer contact visually while controlling closure.
UMI collects demonstrations with a handheld gripper that interacts
directly with the object~\cite{chi2024umi}.

Force and tactile feedback have reduced applied forces and damaging
errors in surgical studies~\cite{wagner2007benefit,king2009tactile}.
Assembly studies show that the benefit depends on the task and
feedback quality~\cite{wildenbeest2013impact}. In a door-opening
imitation-learning task, haptic feedback during demonstration
collection improved data quantity and downstream
performance~\cite{cuan2024haptic}.
These findings support investigating inexpensive feedback for
leader arms. The effectiveness of a servo-based implementation
still needs to be evaluated in its intended task.

FACTR combines leader-arm feedback from follower joint-torque
estimates with gripper-current feedback at the
trigger~\cite{liu2025factr}. Echo uses force-sensing resistors in
the follower fingers to provide grip feedback on a UR teleoperation
interface~\cite{bazhenov2025echo}. AnySkin measures spatially
distributed signals from fingertip deformation using magnetic
sensing~\cite{bhirangi2025anyskin}. We use these signals to detect
contact and activate a trigger position hold, without estimating
force in newtons. Our contribution concerns this contact signal
and rendering mechanism; FACTR has already demonstrated combined
arm and gripper feedback.

\begin{figure*}[t]
  \centering
  \begin{minipage}[b]{0.24\textwidth}
    \centering
    \includegraphics[angle=-90,origin=c,width=\linewidth,height=53mm,
      keepaspectratio]{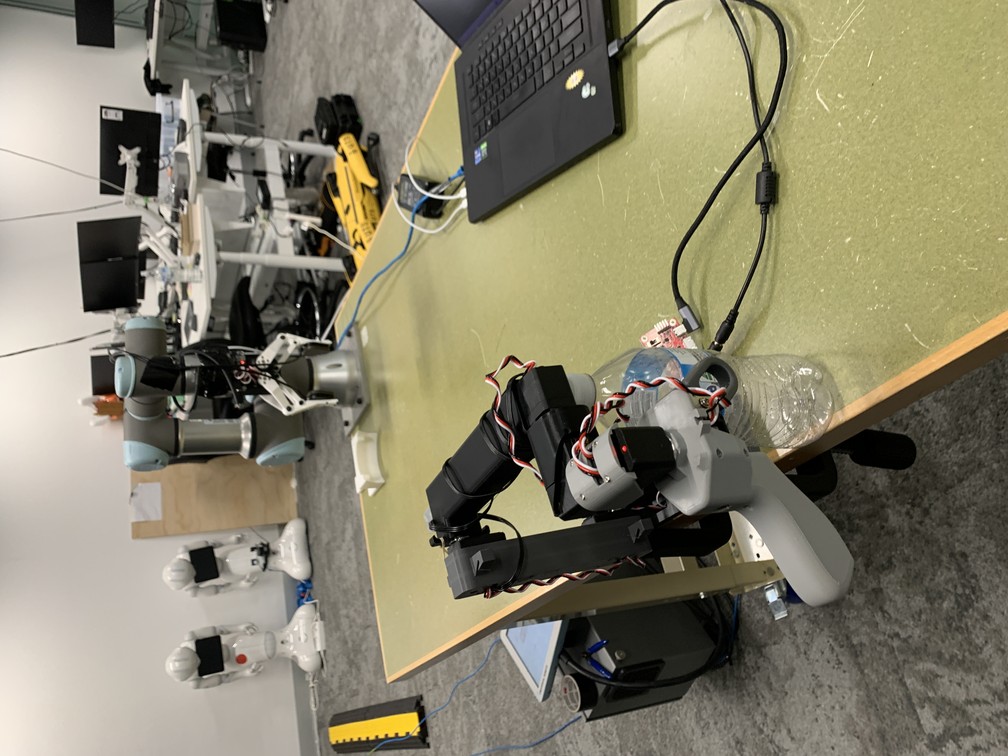}\\[2pt]
    {\footnotesize\textbf{(a)} Leader and follower}
  \end{minipage}\hfill
  \begin{minipage}[b]{0.24\textwidth}
    \centering
    \includegraphics[width=\linewidth,height=53mm,keepaspectratio]{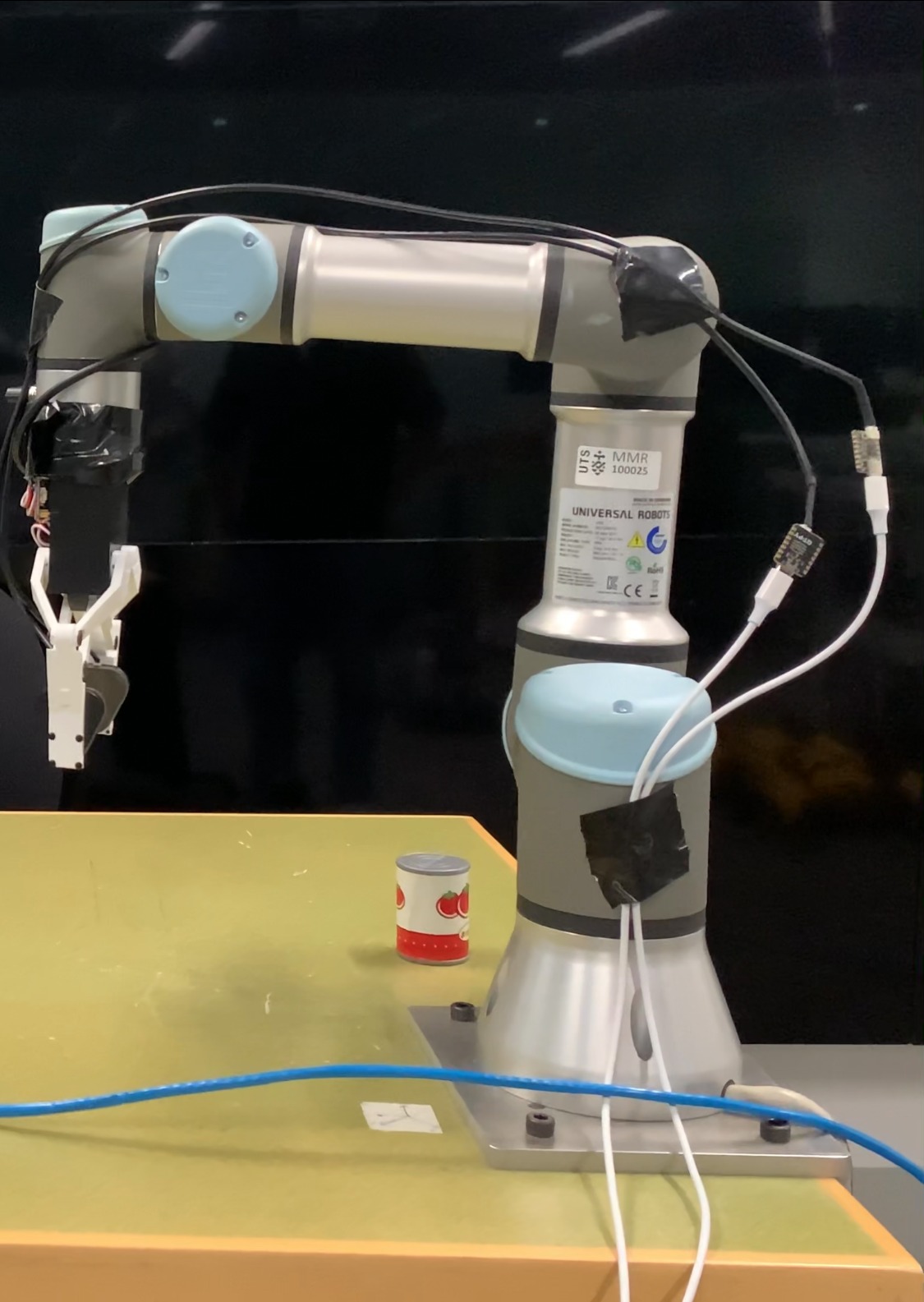}\\[2pt]
    {\footnotesize\textbf{(b)} UR3 and gripper}
  \end{minipage}\hfill
  \begin{minipage}[b]{0.24\textwidth}
    \centering
    \includegraphics[width=\linewidth,height=53mm,keepaspectratio]{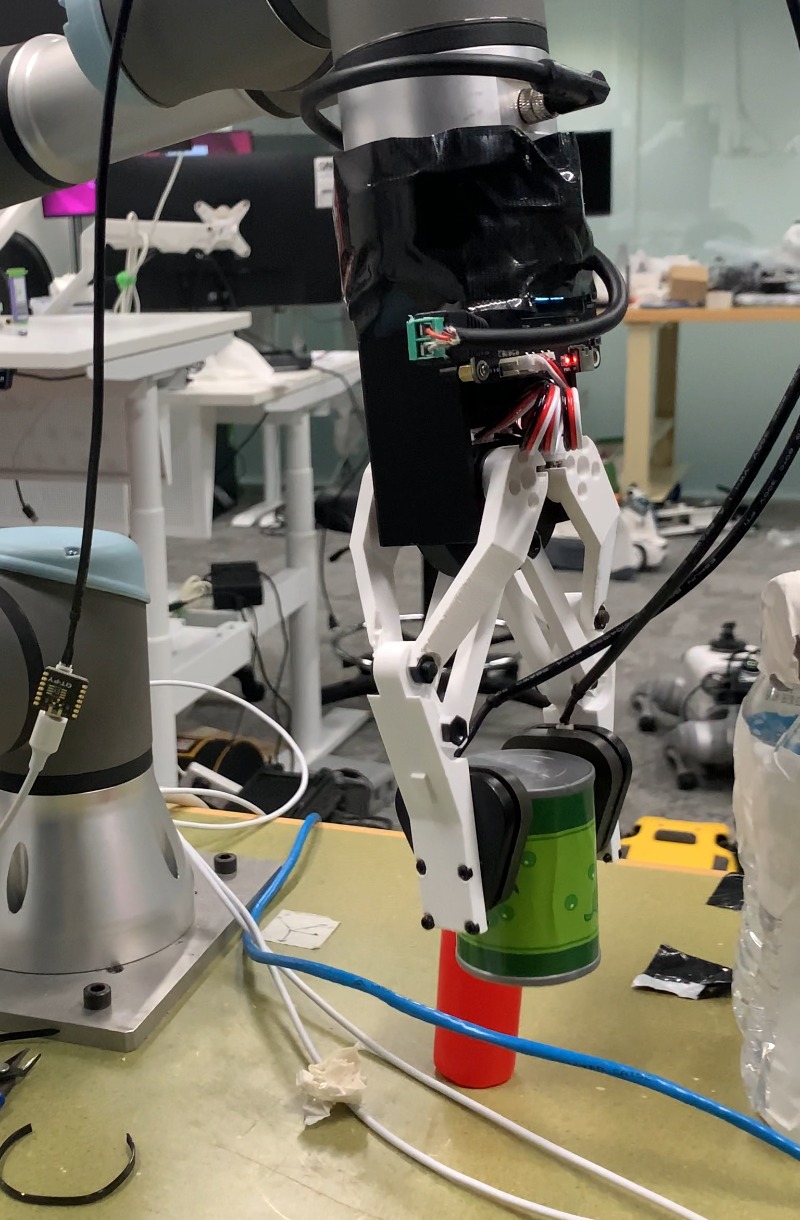}\\[2pt]
    {\footnotesize\textbf{(c)} Container grasp}
  \end{minipage}\hfill
  \begin{minipage}[b]{0.24\textwidth}
    \centering
    \includegraphics[width=\linewidth,height=53mm,keepaspectratio]{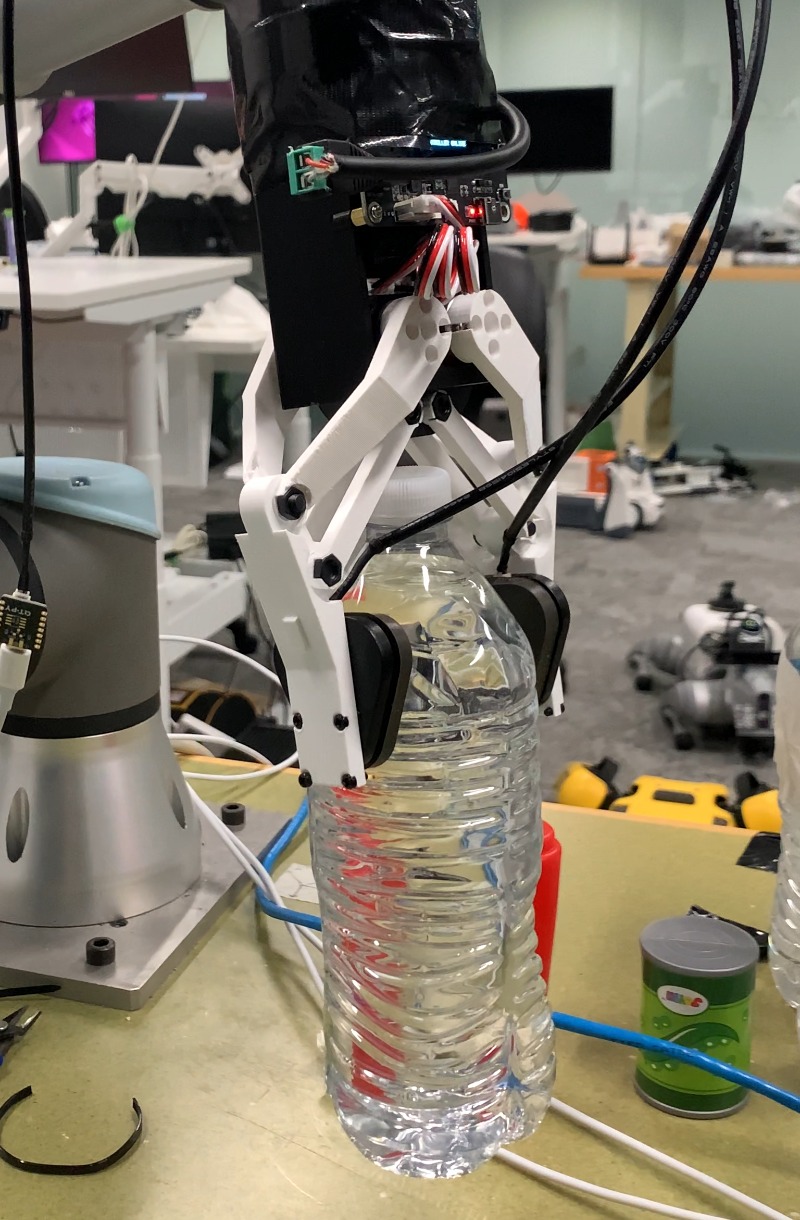}\\[2pt]
    {\footnotesize\textbf{(d)} Bottle grasp}
  \end{minipage}
  \caption{TacGELLO hardware and example grasps. (a) The 3D-printed
  leader in the foreground and the UR3 follower behind it.
  (b) The UR3 fitted with the two-finger gripper.
  (c,d) The gripper holding a cylindrical container and a water bottle.
  Each fingertip carries an AnySkin tactile sensor. The photographs
  are cropped for presentation and illustrate the hardware only;
  they are not used to score experimental outcomes.}
  \label{fig:teaser}
\end{figure*}

TacGELLO is a 3D-printed leader and tactile gripper for the Universal
Robots UR3 (Fig.~\ref{fig:teaser}). It uses existing leader servos for
two feedback channels: tactile contact feedback at the gripper
trigger and current-based load feedback at the arm joints.
Fingertip contact activates a capped hold on the trigger servo.
The load-feedback channel maps changes in UR3 joint current to
output limits on selected leader servos, commanding an arm-resistance
cue during lifting without a dedicated arm force/torque sensor.
This cue has not been calibrated as payload weight or rendered
joint torque. We contribute:
\begin{itemize}
  \item A UR3 leader and two-finger tactile gripper that render
        fingertip contact through an existing trigger servo.
  \item A current-based load-feedback channel that commands
        resistance at selected leader-arm joints from UR3
        joint-current changes.
  \item A characterization of sent-command tracking, selected
        software timings and exploratory fragile-object grasping
        behavior, with arm-load pilots.
\end{itemize}
The grasping sessions compare two complete interfaces with different
command dynamics and contact references. They characterize the
prototype, but cannot isolate the effect of tactile feedback.

\section{Related Work}
\label{sec:related}

\subsection{Low-cost teleoperation interfaces}
GELLO uses kinematically matched leader replicas built from hobby
servos~\cite{wu2024gello}. ALOHA and ALOHA~2 use backdrivable leader
arms for bimanual puppeteering~\cite{zhao2023aloha,aloha2team2024},
and AirExo uses wearable exoskeletons~\cite{fang2024airexo}.
Passive remote leaders transmit the operator's motion without
reflecting follower contact forces. UMI replaces the leader arm
with a handheld gripper~\cite{chi2024umi}. Contact occurs locally
at the device during demonstration collection, so it does not
provide remote follower-force reflection.

\subsection{Force feedback and bilateral control}
Classical bilateral teleoperation established the tradeoff between
transparency and stability~\cite{lawrence1993stability,hokayem2006bilateral}.
Bilateral-control imitation learning uses observer-based force
estimation to exchange position and force information, recording
both for policy learning~\cite{adachi2018imitation,buamanee2024biact,kobayashi2025bilat,yamane2025sensorless}.
FACTR provides compensated leader-arm feedback from follower
external joint torques and gripper feedback from follower gripper
current~\cite{liu2025factr}.
FACTR~2 studies learned external-force estimation on commodity
arms~\cite{oh2026factr2}.
GELLO+F renders Franka external-torque estimates through
current-controlled leader motors~\cite{sujit2025gelloforce}.
Input-gated bilateral teleoperation is another approach for
low-cost hardware~\cite{kanai2025inputgated}.

Echo uses a force-sensing resistor in each follower finger and
renders grip feedback at the leader trigger, with passive arm
joints~\cite{bazhenov2025echo}. Prometheus renders grip-force feedback
in a handheld motion-capture controller~\cite{satsevich2025prometheus}.
TacGELLO uses magnetic fingertip contact sensing to activate a capped
trigger hold, alongside a separate arm-current cue. We have not
compared its effectiveness directly with these systems.

\begin{table}[t]
\centering
\caption{Feedback source and presentation in selected interfaces.
Local interaction in a handheld device differs from remote
follower-force reflection.}
\label{tab:comparison}
\small
\begin{tabular}{@{}lp{3.8cm}@{}}
\toprule
System & Signal and presentation \\
\midrule
GELLO~\cite{wu2024gello} & passive remote leader \\
ALOHA~\cite{zhao2023aloha} & passive remote leader \\
UMI~\cite{chi2024umi} & handheld gripper; local mechanical interaction \\
Bi-ACT~\cite{buamanee2024biact} & observer-based joint-force feedback \\
FACTR~\cite{liu2025factr} & external joint torque to arm; gripper current to trigger \\
GELLO+F~\cite{sujit2025gelloforce} & Franka external torque to current-controlled leader \\
Echo~\cite{bazhenov2025echo} & finger force sensors to trigger; arm passive \\
TactAR~\cite{xue2025rdp} & tactile state as visual AR overlay \\
\textbf{TacGELLO} & tactile contact to trigger hold; joint-current change to arm cue \\
\bottomrule
\end{tabular}
\end{table}

\subsection{Tactile sensing and policy learning}
AnySkin is a replaceable magnetic tactile skin consisting of a
magnetized elastomer over five 3-axis magnetometers, with readings
that respond to normal and shear deformation~\cite{bhirangi2025anyskin}.
It builds on ReSkin~\cite{bhirangi2021reskin}; eFlesh extends magnetic
sensing to customizable 3D-printed geometries~\cite{pattabiraman2025eflesh}.
TactAR presents tactile state visually through augmented reality
alongside Reactive Diffusion Policy~\cite{xue2025rdp}.
DOGlove provides haptic feedback for dexterous
hands~\cite{zhang2025doglove}. TacGELLO presents a contact event at
the trigger of a parallel-jaw leader.

Visual-tactile policies and human contact measurements have been
used for fine-grained manipulation~\cite{huang2024vitac,adeniji2025ftf}.
ForceMimic, FoAR and Adaptive Compliance Policy further motivate
recording force-related information during
demonstrations~\cite{liu2025forcemimic,he2025foar,hou2025acp}.
These studies motivate future policy-learning experiments with
TacGELLO. The present evaluation covers teleoperation only.

\section{System Design}
\label{sec:system}

\begin{figure*}[t]
\centering
\begin{tikzpicture}[font=\footnotesize,
  box/.style={draw, rounded corners=2pt, align=center, text width=22mm,
              minimum height=11mm, inner sep=3pt},
  lab/.style={font=\scriptsize, align=center},
  arr/.style={-{Stealth[length=1.8mm]}, semithick},
  fb/.style={arr, draw=blue!70!black}]
  \node[box, minimum height=32mm] (pc)
    {\textbf{Host PC}\\[2pt] align \& smooth leader angles\\[1pt]
     control interval ${\approx}17$\,ms\\[1pt] contact cue\\[1pt] load cue\\[1pt] logging};
  \node[box, anchor=east, minimum height=20mm] (leader) at ([xshift=-36mm]pc.west)
    {\textbf{Leader arm}\\[2pt] 3D-printed UR3 replica\\[1pt] 6 joint servos\\[1pt] trigger servo};
  \node[box, anchor=west] (ur) at ([xshift=40mm, yshift=9mm]pc.east)
    {\textbf{UR3}\\ RTDE interface};
  \node[box, anchor=west] (grip) at ([xshift=40mm, yshift=-9mm]pc.east)
    {\textbf{Gripper}\\ ESP32, 2 servos\\ AnySkin $\times$2};
  \coordinate (l1) at ([yshift=4mm]leader.east);  \coordinate (p1) at (l1 -| pc.west);
  \coordinate (l2) at ([yshift=-4mm]leader.east); \coordinate (p2) at (l2 -| pc.west);
  \draw[arr] (l1) -- node[above, lab]{joint angles,\\trigger position} (p1);
  \draw[fb]  (p2) -- node[below, lab]{trigger hold (contact),\\joint output limits (load)} (l2);
  \coordinate (u1) at ([yshift=2mm]ur.west);  \coordinate (q1) at (u1 -| pc.east);
  \coordinate (u2) at ([yshift=-2mm]ur.west); \coordinate (q2) at (u2 -| pc.east);
  \draw[arr] (q1) -- node[above, lab]{servoJ targets} (u1);
  \draw[fb]  (u2) -- node[below, lab]{joint angles, currents} (q2);
  \coordinate (g1) at ([yshift=2mm]grip.west);  \coordinate (r1) at (g1 -| pc.east);
  \coordinate (g2) at ([yshift=-2mm]grip.west); \coordinate (r2) at (g2 -| pc.east);
  \draw[arr] (r1) -- node[above, lab]{width commands via\\TCP/Bluetooth bridge} (g1);
  \draw[fb]  (g2) -- node[below, lab]{AnySkin via USB;\\finger positions via Bluetooth} (r2);
\end{tikzpicture}
\caption{Experimental system architecture. Black: command path. Blue: signals used
for feedback. Feedback commands act through the leader's own servos,
by holding the trigger and adjusting joint servo output limits.
The AnySkin USB streams are separate from the ESP32 motor-control link.
No dedicated arm force/torque sensor or additional leader actuator is used.}
\label{fig:architecture}
\end{figure*}
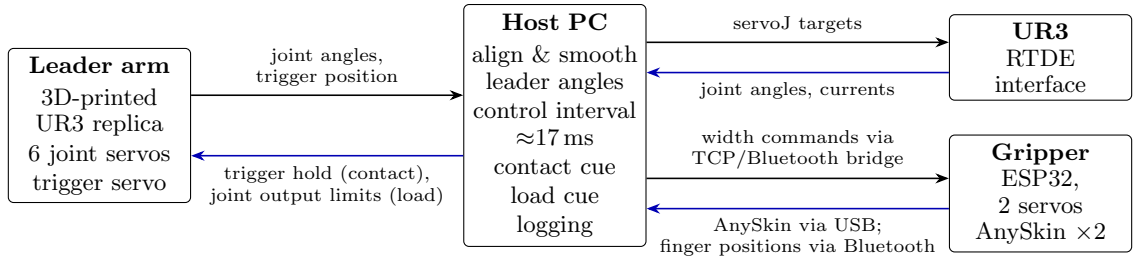

\subsection{Hardware}
The leader is a GELLO-style~\cite{wu2024gello} 3D-printed,
kinematically matched replica of the UR3, with six joint servos and
a trigger servo, all STS3215 serial-bus actuators. We enable the
leader servos with limited output to render the cues.

We designed a 3D-printed two-finger gripper for the UR3
(Fig.~\ref{fig:teaser}b--d). Two STS3215 serial-bus servos drive its
printed linkage fingers through an ESP32 controller, over a commanded
opening range of 0--130\,mm. An AnySkin skin~\cite{bhirangi2025anyskin}
forms the contact pad of each fingertip, so the surface that grasps
the object is also the surface that senses contact. Each sensor's
magnetometers are read by its own Adafruit QT~Py microcontroller,
which streams the readings to the host over USB. AnySkin
separates the replaceable skin from its sensing electronics, so a
worn pad can be swapped without rewiring the gripper.

\subsection{Teleoperation loop}
\label{sec:loop}
The host PC reads leader joint angles, applies a fixed startup
alignment offset and smooths the targets before streaming commands
to the UR3 through the RTDE servoJ interface
(Fig.~\ref{fig:architecture}). The median valid control interval
in the grasping recordings is approximately 16.65\,ms.
The host maps trigger position to commanded gripper width and
publishes a ROS command. A host-side TCP/Bluetooth bridge forwards
motor commands to the ESP32, which drives the gripper servos over
their serial bus. The AnySkin measurements return separately over
USB. Section~\ref{sec:exp_grasp} reports the empirical
trigger-to-width calibration for the two interfaces.
We timestamp host events with a monotonic clock. RTDE supplies
robot timestamps, but the two clocks are not synchronized, which
limits the timing measurements.

\subsection{Fingertip channel: contact at the trigger}
\label{sec:tactile}
Each AnySkin fingertip contains five 3-axis magnetometers. Contact
detection compares field magnitudes with an empty-gripper baseline
that varies with gripper configuration, without estimating force
in newtons. We record each fingertip's five field magnitudes,
baseline values, grouped deviations and controller contact flag,
with a detection margin of 20 sensor units. The grouped deviations
keep the first and last magnetometer differences and the maximum
of the three middle differences. The controller contact flag drives
the trigger hold and defines contact onset in the analysis. Because
the flag also reflects gripper operation and release, a threshold
applied independently to the stored deviations would define a
different contact reference.

On contact, the existing trigger servo holds its measured position
to provide a contact-resistance cue. The experimental setup
specified an output limit of 150 raw servo units. The gripper
controller writes a position target and enables servo output,
then releases the hold when the trigger moves toward open or
contact clears. It also
attempts release when contact messages become stale. Output-limit
readback was not recorded, so the cap was not independently verified.
Resisting torque and the force needed to override the hold were
not calibrated.
The cue conveys contact onset, with no graded grip-force estimate.

\subsection{Arm channel: current-driven resistance cue}
\label{sec:load}
Measured UR3 shoulder, elbow and first-wrist joint currents are
compared with an empty-arm baseline.
The arm cue uses position-lag control with limited servo output.
Changes in follower current adjust the output limits of the
corresponding leader servos to command resistance during lifting.
Section~\ref{sec:exp_load} gives the settings for each load condition.
Posture and motion also affect current, so this signal is not a
calibrated weight estimate. We evaluate cue-command activation;
rendered joint torque has not been measured.

\subsection{Operating limits}
The teleoperation code limits joint-command steps and checks encoder
ranges. Leader-servo output limits have not been calibrated as
physical torque or manual-override force. Cue release after host
failure or communication loss has not been tested.
We tested non-ferromagnetic
objects because ferromagnetic material can affect magnetic sensing.

\subsection{Adapting the interface}
Adapting the fingertip cue requires a compatible follower-gripper
interface. The arm cue also needs a joint mapping, appropriate
current or torque measurements and control settings for that
follower. A kinematic leader replica must match the new arm.
We have not evaluated portability beyond the UR3.

\section{Experiments}
\label{sec:experiments}

We recorded (E1)~joint-command tracking and selected software
timings, (E2)~fragile-object grasping with two gripper interfaces,
and (E3)~arm-current cue activation during lifting. One author
operated the leader throughout. We collected system logs without
subjective ratings.

\subsection{E1: Command tracking and timing}
\label{sec:exp_tracking}
We compare sent follower joint commands with measured UR3 angles
across the six E2 grasping sessions, using the host-observed sample
associations. The sent commands include leader
offset alignment and smoothing. This measures follower command
tracking; the complete error between raw leader measurements and
follower motion is outside this measurement.

We pool samples across all six sessions to compute same-time RMSE.
For each joint-session with at least $30^\circ$ of cumulative
commanded travel, we linearly interpolate both angle trajectories
onto a 2\,ms grid and take first differences. We estimate lag as
the nonnegative shift below 400\,ms that maximizes the normalized
dot product of those differences, then report the median lag and
median aligned RMSE over qualifying sessions. The 2\,ms interpolation
grid does not give physical timestamp accuracy at that scale.
The robot and host clocks are not synchronized.

For each software stage, we take the median of available session
medians: six sessions for arm stages and the three TacGELLO sessions
for gripper stages. Effective recording rate is the median, across sessions,
of sample count divided by recording duration. We also report the
median midpoint of 634 host-observed motion-response brackets for
$0.1^\circ$ displacement. Transport, polling and continuous command
streaming prevent assigning this statistic to the latency of an
individual command.

\subsection{E2: Fragile-object grasping}
\label{sec:exp_grasp}
The operator repeatedly grasped and released a potato chip, an empty
paper cup and a thin-walled plastic bottle. Each object was replaced
between grasps. For each object type, we recorded one TacGELLO
session with direct trigger-to-width commands and tactile trigger
hold, and one original-interface session with feedback off and
rate-limited commands. TacGELLO was recorded first for the chip and
bottle; the original interface was first for the cup. Conditions
were not randomized or blinded.

Both feedback and command dynamics change between conditions.
During closing samples, median command deviation from the common
nominal trigger-to-width target was 16--27\,mm for the original
interface and approximately 0.1\,mm for TacGELLO. Median
original-interface closing speeds were 41--44\,mm/s. These checks
describe the recorded command behavior, without establishing
identical mapping calibration across conditions.

\paragraph{Segmentation and contact references.}
We segment consecutive samples whose commanded width falls below
0.8 times the session's 95th-percentile commanded width, retaining
runs with a first-to-last-sample duration exceeding 0.5\,s.
Thresholds were 104.0\,mm for all TacGELLO sessions and 99.8, 101.9
and 101.7\,mm for the original chip, cup and bottle sessions.
Table~\ref{tab:counts} gives interval counts. Grasp success was
not independently verified.

For each interval, we use the first logged tactile onset from
0.5\,s before its start to its end. If no tactile onset is
available, we use the first commanded-width sample in that window
at or below the median commanded width at all tactile onsets in
the corresponding TacGELLO session: 61.3, 87.2 and 92.4\,mm for
chip, cup and bottle. This width fallback applies to both original
sessions and two TacGELLO intervals. Nine original-interface
intervals never reach the reference and lack contact-dependent
outcomes. Original-interface tactile data were not logged, so
the width reference is a proxy rather than measured physical contact.

\begin{table}[t]
\centering
\caption{Segmented intervals and available contact references.
T: tactile onset; W: width fallback; --: no reference.}
\label{tab:counts}
\small
\setlength{\tabcolsep}{3.5pt}
\begin{tabular}{@{}llrrrr@{}}
\toprule
Object & Interface & Total & T & W & -- \\
\midrule
Chip & TacGELLO & 19 & 18 & 1 & 0 \\
     & Original & 23 & 0 & 21 & 2 \\
Cup & TacGELLO & 20 & 19 & 1 & 0 \\
    & Original & 17 & 0 & 14 & 3 \\
Bottle & TacGELLO & 23 & 23 & 0 & 0 \\
       & Original & 25 & 0 & 21 & 4 \\
\bottomrule
\end{tabular}
\end{table}

\paragraph{Outcome definitions.}
We compute commanded closure after the contact reference as the
nonnegative difference between width at the reference and minimum
commanded width from that reference to 0.2\,s after the segment end.
This is a command change; physical aperture and force are unmeasured.
Trigger rotation uses the minimum valid encoder value from 1.5\,s
before the segment start to its end, relative to a common engagement
reference of 3924.712 counts. Samples at or below 3000 counts are
rejected as wrap artifacts. We use $360/4096$ degrees per count.

Finger-servo closure is the average of the two servos' individual
maximum closing rotations relative to their fully open references,
using samples from 1.5\,s before the segment start to 0.5\,s after
its end. The two extrema need not occur simultaneously. Open
references are medians while commanded width is at least 129.5\,mm.
One original bottle interval lacks the necessary finger samples.
These encoder measures are not calibrated grip force or damage.

Five steady-state trigger-to-width fits have slopes of
0.234--0.237\,mm/count. The original bottle fit is anomalous:
0.123\,mm/count with 12.43\,mm residual RMSE. We therefore cannot
establish equal empirical calibration. Trigger results retain the
common reference above, with this limitation.

We report medians, quartiles and Cliff's $\delta$, where negative
$\delta$ indicates smaller TacGELLO values. Repeated intervals share
an operator and session. With one session per object-condition,
per-grasp significance tests cannot establish a causal or
population-level feedback effect, so we use descriptive summaries.

\subsection{E3: Arm-current feedback during lifting}
\label{sec:exp_load}

We performed preliminary lifting tests to record changes in
UR3 joint current and the corresponding resistance commands
generated for the leader arm.

Two conditions were tested. In the first, the operator alternated
between a 20\,g toy can and a 600\,mL water bottle, performing
two grasps with each object. A 1200\,mL water bottle was tested
separately.

The toy-can and 600\,mL bottle condition used a 0.1\,A deadband,
a 0.24\,A current range and gain 0.6 applied to a cap of 40,
giving a maximum commanded output limit of 24 raw servo units.
The 1200\,mL bottle condition used a 0.20\,A current range and
a cap of 40 raw units; its gain was not recorded. Results are
analyzed separately because the controller settings differed.

We recorded follower joint currents, the empty-arm baseline,
commanded leader-servo output limits and the feedback-active
flag. A cue-activation interval is a continuous period during
which this flag is active. The first condition contained three
such intervals and the 1200\,mL condition contained seven.

For each interval, we calculate the median signed shoulder-current
change relative to the empty-arm baseline using valid active
samples. The interval includes its start and excludes the
cue-inactive sample at its end.

These measurements characterize feedback-command activation.
Rendered joint torque and operator perception were not measured.
Results are reported in Sec.~\ref{sec:res_load}.
\section{Results}
\label{sec:results}

\subsection{E1: Command tracking and timing}
\label{sec:res_tracking}

\begin{figure}[t]
\centering
\includegraphics[width=\linewidth]{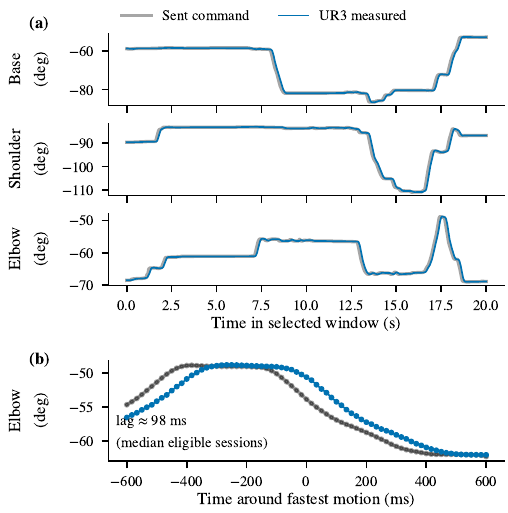}
\caption{Sent joint commands and measured UR3 angles.
(a)~Three joints during the selected 20\,s window with the most
commanded travel. (b)~Zoom on the fastest motion of the most active
joint in that window. The displayed command has already undergone
leader offset alignment and smoothing.}
\label{fig:tracking}
\end{figure}

\begin{table}[t]
\centering
\caption{Sent-command tracking over six sessions. Same-time RMSE
pools samples; lag and aligned RMSE are medians over eligible
joint-sessions ($n$). Software rows use available session medians:
six for arm stages and three for gripper stages.}
\label{tab:tracking}
\small
\begin{tabular}{@{}lcccc@{}}
\toprule
Joint & Same-time & Lag & Aligned & $n$ \\
 & RMSE ($^\circ$) & (ms) & RMSE ($^\circ$) & \\
\midrule
Base & 0.18 & 84 & 0.014 & 3 \\
Shoulder & 0.19 & 98 & 0.017 & 3 \\
Elbow & 0.36 & 98 & 0.028 & 5 \\
Wrist 1 & 0.27 & 98 & 0.028 & 4 \\
Wrist 2 & 0.15 & 84 & 0.023 & 5 \\
Wrist 3 & 0.16 & 82 & 0.017 & 5 \\
\midrule
\multicolumn{3}{@{}l}{Effective recording rate} & \multicolumn{2}{r@{}}{57.6\,Hz} \\
\multicolumn{3}{@{}l}{Leader read} & \multicolumn{2}{r@{}}{1.40\,ms} \\
\multicolumn{3}{@{}l}{Read end $\rightarrow$ servoJ return} & \multicolumn{2}{r@{}}{0.29\,ms} \\
\multicolumn{3}{@{}l}{Gripper receive $\rightarrow$ send} & \multicolumn{2}{r@{}}{13.4\,ms} \\
\multicolumn{3}{@{}l}{Gripper send $\rightarrow$ ACK} & \multicolumn{2}{r@{}}{10.1\,ms} \\
\multicolumn{3}{@{}l}{Motion-bracket midpoint ($n{=}634$)} & \multicolumn{2}{r@{}}{56.6\,ms} \\
\bottomrule
\end{tabular}
\end{table}

Same-time sent-command tracking error was 0.15--0.36$^\circ$ RMSE.
After alignment by an estimated 82--98\,ms trajectory lag,
residual RMSE was 0.014--0.028$^\circ$
(Fig.~\ref{fig:tracking}, Table~\ref{tab:tracking}).
Alignment measures agreement between interpolated command and
measured trajectories. Physical leader accuracy has not been
established at that scale. Median valid consecutive control
intervals were approximately 16.65\,ms, and the median
recording-level rate was 57.6\,Hz.

Median session-level leader-read and read-to-servoJ-return durations
were 1.40 and 0.29\,ms. These timings cover selected arm-command
execution stages, leaving complete end-to-end latency unmeasured.
The median host-observed motion-bracket midpoint was 56.6\,ms,
with median lower and upper endpoints of 48.3 and 65.0\,ms. This
midpoint is not an upper bound on an isolated command's response. Gripper
message-receipt-to-send and send-to-acknowledgement stage medians
were 13.4 and 10.1\,ms. The logs cannot separate the contributions
of controller dynamics, smoothing, transport and sampling to
trajectory lag. Contact-to-trigger actuation latency was also
unmeasured.

\subsection{E2: Fragile-object grasping}
\label{sec:res_grasp}

\begin{figure}[t]
\centering
\includegraphics[width=\linewidth]{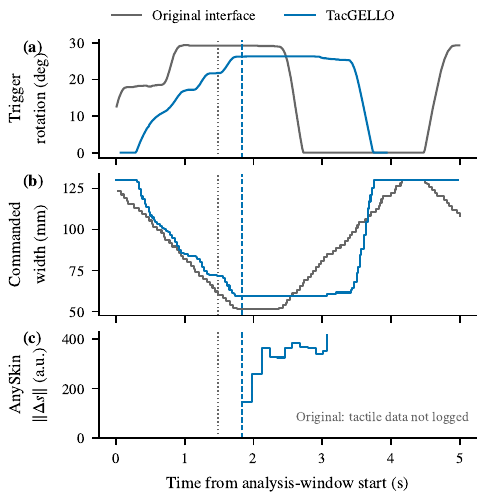}
\caption{Representative chip intervals selected near each interface's
median minimum commanded width, with pre-window commanded width
at least 125\,mm and segment duration under 6\,s. Traces are aligned to 0.6\,s before each
segmented interval. The TacGELLO marker is logged tactile onset;
the original-interface marker is its width-based contact reference.
The AnySkin trace is the norm of baseline-relative logged field
magnitudes and has no force calibration.}
\label{fig:grasp_trace}
\end{figure}

\begin{figure}[t]
\centering
\includegraphics[width=\linewidth]{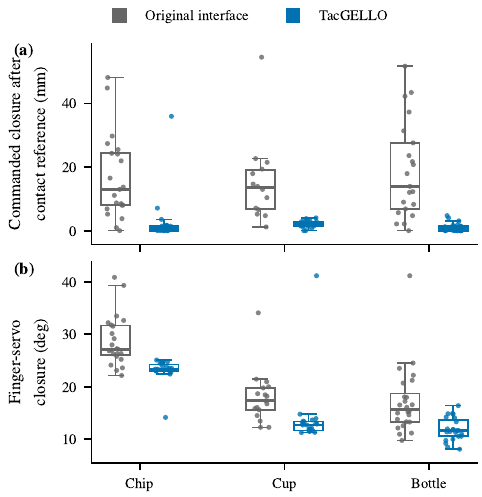}
\caption{Per-interval command and encoder measurements for the
original interface (grey) and TacGELLO (blue). Contact-dependent
values use tactile onset where available and a width reference
otherwise, including two TacGELLO fallbacks. Conditions also differ
in command dynamics. Points include the large chip outlier.}
\label{fig:grasp_comparison}
\end{figure}

\begin{table*}[t]
\centering
\caption{Exploratory grasp outcomes: median [Q1, Q3]. Original:
feedback off, rate-limited commands; TacGELLO: tactile hold, direct
commands. Outcome-specific sample sizes are given in each row.
$\delta$ is Cliff's delta for TacGELLO minus original. Commanded
closure uses different contact references across conditions.}
\label{tab:grasp}
\small
\setlength{\tabcolsep}{5pt}
\begin{tabular}{@{}llccccc@{}}
\toprule
Object & Measure & Original & $n$ & TacGELLO & $n$ & $\delta$ \\
\midrule
Chip & Commanded closure (mm) & 13.0 [8.0, 24.3] & 21 & 0.7 [0.0, 1.5] & 19 & $-0.77$ \\
     & Trigger rotation ($^\circ$) & 31.0 [28.7, 35.9] & 23 & 26.6 [25.8, 27.9] & 19 & $-0.62$ \\
     & Finger-servo closure ($^\circ$) & 27.2 [26.0, 31.7] & 23 & 23.4 [23.0, 24.2] & 19 & $-0.81$ \\
\midrule
Cup & Commanded closure (mm) & 13.5 [6.7, 19.0] & 14 & 2.1 [1.4, 2.9] & 20 & $-0.88$ \\
    & Trigger rotation ($^\circ$) & 20.0 [18.2, 23.0] & 17 & 16.9 [15.8, 18.3] & 20 & $-0.49$ \\
    & Finger-servo closure ($^\circ$) & 17.3 [15.6, 19.8] & 17 & 12.7 [11.7, 13.3] & 20 & $-0.76$ \\
\midrule
Bottle & Commanded closure (mm) & 13.9 [6.8, 27.6] & 21 & 0.7 [0.0, 1.4] & 23 & $-0.90$ \\
       & Trigger rotation ($^\circ$) & 17.7 [15.3, 20.8] & 25 & 14.2 [13.6, 16.9] & 23 & $-0.41$ \\
       & Finger-servo closure ($^\circ$) & 15.7 [13.3, 18.7] & 24 & 11.6 [10.5, 13.7] & 23 & $-0.67$ \\
\bottomrule
\end{tabular}
\end{table*}

\paragraph{Contact registration and hold behavior.}
We found logged tactile onset within the association window for
60 of 62 segmented TacGELLO intervals. The other two used width
references in Table~\ref{tab:grasp}. The cup's first interval began
already below the segmentation threshold. Without independent
contact labels, we cannot classify it as a confirmed missed contact.
One additional tactile-onset event occurred outside segmentation
windows. Its status as a
physical contact or false positive is unverified.

In 59 of the 60 tactile-referenced intervals, subsequent commanded
closure was below 5\,mm. The exception was a chip interval with
35.9\,mm of subsequent commanded closure and duration 17.6\,s.
We retain this interval in the results; its cause is unknown from
the logs. Across all 62 TacGELLO intervals, including the two width
fallbacks, median commanded closure after the reference was 0.7,
2.1 and 0.7\,mm for chip, cup and bottle.

\paragraph{Comparison of the complete interfaces.}
Original-interface median commanded closure after its width reference
was 13.0, 13.5 and 13.9\,mm. Peak trigger rotation and finger-servo
closure were also larger in the original sessions
(Table~\ref{tab:grasp}). These are measurements of commands and
encoder motion. Grip force, object deformation, damage and
successful lifting were not directly measured.

In the representative trace, the original command trails the trigger
and TacGELLO holds near the tactile reference
(Fig.~\ref{fig:grasp_trace}). Command dynamics, contact references
and session conditions differ between interfaces. The reported
effect sizes describe those recordings, so the reduction cannot
be attributed to tactile feedback alone.

\subsection{E3: Arm-current feedback during lifting}
\label{sec:res_load}

We summarize the recorded shoulder-current changes and
leader-servo output-limit commands for the two E3 conditions.
These measurements characterize feedback-command activation;
rendered joint torque and operator perception were not measured.

The alternating toy-can and 600\,mL bottle condition contained
three cue-activation intervals. Median signed shoulder-current
changes relative to the empty-arm baseline were 0.528, 0.407
and 0.138\,A, calculated from 123, 73 and 141 active samples,
respectively. Peak commanded output limits were 24, 24 and
14 raw units. The configured maximum was reached in two
intervals. Individual intervals were not reliably associated
with the toy can or bottle, so these values are reported at
the recording level.

The separate recording associated with the nominal 1200\,mL
bottle condition contained seven cue-activation intervals,
comprising 286 active samples (Table~\ref{tab:load1200}).
Median signed shoulder-current changes ranged from 0.110 to
0.572\,A. Peak commanded output limits ranged from 4 to
40 raw units, reaching the configured cap in one interval.

The intervals identify periods of controller activity and do
not establish the number of successful lifts. Bottle volume
and total payload mass were not independently verified.
Different controller settings and uncontrolled posture also
prevent a matched comparison between payloads. The results
therefore demonstrate recorded feedback-command activation
without establishing load-proportional resistance or
perceptible weight feedback.

\begin{table}[t]
\centering
\caption{Logged cue activations associated with the nominal
1200\,mL bottle condition. $n$ counts valid active samples.
$\Delta I$ is the median signed shoulder-current change
relative to the empty-arm baseline. Peak limits are logged
output-limit commands in raw servo units.}
\label{tab:load1200}
\small
\setlength{\tabcolsep}{4pt}
\begin{tabular}{@{}rrrr@{}}
\toprule
Cue & $n$ & $\Delta I$ (A) & Peak limit \\
\midrule
1 & 38  & 0.572 & 27 \\
2 & 25  & 0.287 & 10 \\
3 & 134 & 0.280 & 40 \\
4 & 29  & 0.422 & 29 \\
5 & 25  & 0.144 & 10 \\
6 & 9   & 0.110 & 5  \\
7 & 26  & 0.158 & 4  \\
\bottomrule
\end{tabular}
\end{table}

\section{Discussion and Limitations}
\label{sec:discussion}

TacGELLO uses existing leader servos to command fingertip-contact
and arm-load cues. The grasping recordings showed lower commanded
closure with TacGELLO, but feedback, command dynamics and contact
references changed together. One operator performed one session
per object-condition without randomized ordering. The original
bottle calibration anomaly adds uncertainty to trigger comparisons.
These differences prevent attributing the observed reduction
to tactile feedback alone.

The measurements characterize commands and encoder motion.
Physical-contact accuracy, grasp outcomes, rendered joint torque
and operator perception were not independently evaluated.
Joint current also depends on posture and motion, and the load
tests used different controller settings without independently
verified payload masses. The arm results therefore support
feedback-command activation, while load-proportional rendering
remains unverified. Similarly, estimated trajectory lag does not
measure complete contact-to-trigger latency.

A matched comparison should keep command dynamics and contact
measurement identical while varying the trigger hold across
randomized trials with additional operators. A contact-stop
condition without trigger resistance would help distinguish
automatic stopping from the contribution of the operator cue.
Arm-feedback validation requires known masses at controlled
poses, torque calibration and operator assessments. Complete
reproduction also requires the detector filtering settings and
experimental arm-controller source. Evaluation remains limited
to the tested UR3 setup and non-ferromagnetic objects.
\section{Conclusion}
\label{sec:conclusion}

We built TacGELLO to return fingertip contact to a 3D-printed UR3
leader through its existing trigger servo. A separate channel uses
follower joint-current changes to provide an arm-resistance cue.
Recordings from one operator characterize sent-command tracking
and show lower commanded closure in a preliminary comparison with
the original gripper interface. Command dynamics, contact
references and session conditions also differed, leaving the effect
of tactile feedback unresolved. A matched comparison with independent
contact and grasp outcomes, together with calibration of the rendered
cues, is needed to evaluate the mechanism's effectiveness.

\newpage
\newpage
\bibliographystyle{named}
\bibliography{references}
\end{document}